\documentclass[%
 reprint,
superscriptaddress,
nofootinbib,
 amsmath,amssymb,
 aps,
 longbibliography
a4paper
]{revtex4-1}

\usepackage{graphicx}
\usepackage{dcolumn}
\usepackage{bm}
\usepackage{array}
\usepackage{dcolumn}
\usepackage{bm}
\usepackage{gensymb}
\usepackage{graphicx}
\usepackage{color}
\usepackage{hyperref}
\usepackage{fontawesome}
\usepackage{url}
\usepackage{soul}
\usepackage[toc,page]{appendix}
\usepackage[utf8]{inputenc}
\usepackage{todonotes}
\usepackage{lineno}
\usepackage{todonotes}
\usepackage{tikz}
\usetikzlibrary{arrows.meta}
\tikzset{%
  >={Latex[width=2mm,length=2mm]},
}

\makeatletter
\newcommand{\thickhline}{%
    \noalign {\ifnum 0=`}\fi \hrule height 1pt
    \futurelet \reserved@a \@xhline
}
\newcolumntype{M}[1]{>{\centering\arraybackslash}m{#1}}
\newcolumntype{'}{@{\hskip\tabcolsep\vrule width 1pt\hskip\tabcolsep}}

\makeatother

\begin{document}

\title{Constraining the Parameters of High-Dimensional Models with Active Learning}

\author{Sascha Caron}
\affiliation{Institute for Mathematics, Astro- and Particle Physics IMAPP, Radboud Universiteit, Nijmegen, The Netherlands}
\affiliation{Nikhef, Amsterdam, The Netherlands}
\author{Tom Heskes}
\affiliation{Data Science Institute for Computing and Information Sciences (iCIS), Radboud University, Nijmegen, The Netherlands}
\author{Sydney Otten}
\affiliation{Institute for Mathematics, Astro- and Particle Physics IMAPP, Radboud Universiteit, Nijmegen, The Netherlands}
\affiliation{GRAPPA, University of Amsterdam, The Netherlands}
\author{Bob Stienen} \email{b.stienen@science.ru.nl}
\affiliation{Institute for Mathematics, Astro- and Particle Physics IMAPP, Radboud Universiteit, Nijmegen, The Netherlands}

\date{\today}

\begin{abstract}
Constraining the parameters of  physical models with $>5-10$ parameters is a widespread problem in fields like particle physics and astronomy. The generation of data to explore this parameter space often requires large amounts of computational resources. The commonly used solution of reducing the number of relevant physical parameters hampers the generality of the results. 
In this paper we show that this problem can be alleviated by the use of active learning. We illustrate this with examples from high energy physics, a field where simulations are often expensive and parameter spaces are high-dimensional. We show that the active learning techniques query-by-committee and query-by-dropout-committee allow for the identification of model points in interesting regions of high-dimensional parameter spaces (e.g. around decision boundaries). This makes it possible to constrain model parameters more efficiently than is currently done with the most common sampling algorithms and to train better performing machine learning models on the same amount of data. Code implementing the experiments in this paper can be found on GitHub \href{https://github.com/bstienen/active-learning}{\faGithub}.
\end{abstract}
\maketitle
\section{Introduction}
\label{sec:introduction}
With the rise of computational power over the last decades, science has gained the power to evaluate predictions of new theories and models at unprecedented speeds. Determining the output or predictions of a model given a set of input parameters often boils down to running a program and waiting for it to finish. The same is however not true for the inverse problem: determining which (ranges of) input parameters a model can take to produce a certain output (e.g., finding which input parameters of a universe simulation yield a universe that looks like ours) is still a challenging problem. In fields like high energy physics and astronomy, where models are often high-dimensional, determining which model parameter sets are still allowed given experimental data is a time-consuming process that is currently often approached by looking only at lower-dimensional simplified models. This not only requires large amounts of computational resources, in general it also reduces the range of possible physics the model is able to explain.

\begin{figure}
    \begin{center}
        \includegraphics[width=0.375\textwidth]{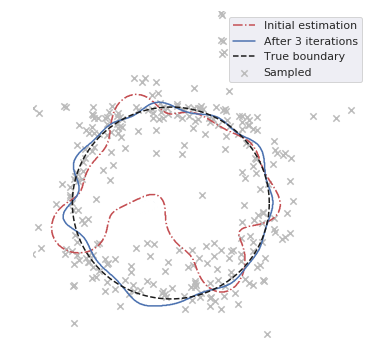}
    \end{center}
    \caption{With active learning new data points can be sampled in regions of interest, like for example a decision boundary in a classification problem. The figure shows how the initial estimation (dashed-dotted red line) of the decision boundary (dashed black line) is located at the location where the classification of new points is most uncertain. By iteratively sampling new points (crosses) in this most uncertain region and determining a new estimation of the decision boundary, the estimation of the boundary will get increasingly more accurate, as can be seen in the picture for 3 iterations (solid blue line). }
    \label{fig:basic_idea}
\end{figure}

In this paper we approach this problem by exploring the use of \textit{active learning}~\cite{settles:2010, Seung:1992:QC:130385.130417, Cohn1994}, an iterative method that applies machine learning to guide the sampling of new model points to specific regions of the parameter space. Active learning reduces the time needed to run expensive simulations by evaluating points that are expected to lie in regions of interest. As this is done iteratively, this method increases the resolution of the true boundary with each iteration. For classification problems this results in the sampling of points around -- and thereby a better resolution on -- decision boundaries. An example of this can be seen in Figure~\ref{fig:basic_idea}. In this paper we investigate techniques called \textit{query-by-committee}~\cite{Seung:1992:QC:130385.130417} and \textit{query-by-dropout-committee}~\cite{dropout_based_al,2015arXiv151106412D, 2018arXiv181103897P}, which allow for usage of active learning in parameter spaces with a high dimensionality.

The paper is structured as follows: in Section~\ref{sec:activelearning} we explain how active learning works. In Section~\ref{sec:experiments} we show applications of active learning to determine decision bounds of a model in the context of high energy physics, working in model spaces of a 19-dimensional supersymmetry (SUSY) model\footnote{Supersymmetry (SUSY) is a theory that extends the current theory of particles and particle interactions by adding another space-time symmetry. It predicts the existence of new particles which could be measured in particle physics experiments, if supersymmetry is realised in nature.}. We conclude the paper in Section~\ref{sec:conclusion} with a summary and future research directions.
\section{Active Learning}
\label{sec:activelearning}

Scientific simulations  can be computationally expensive to run, making it expensive to explore the output space of these. Approximations of these simulations can however be constructed in the form of machine learning estimators, which are typically quick to evaluate. Active learning leverages this speed, exploiting the ability to quickly estimate how much information can be gained by querying a specific point to the true simulation (also known as the labelling procedure).

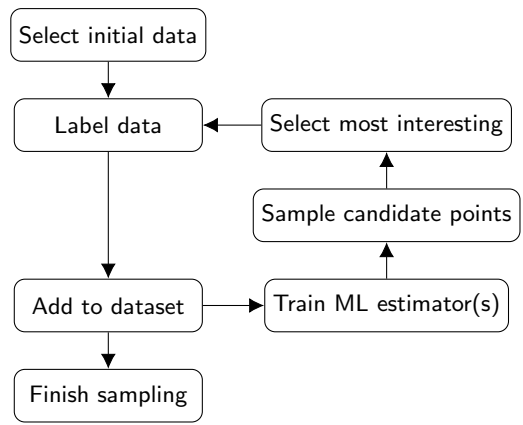
\begin{figure}
	\begin{center}
        \begin{tikzpicture}[node distance=1.2cm, every node/.style={fill=white, font=\sffamily, rectangle, rounded corners, draw=black, minimum width=2.5cm, minimum height=0.7cm, text centered, font=\sffamily}, align=center]
            \node (start)								{\small Select initial data};
            \node (sample)		[below of=start]		{\small Label data};
            \node (dataset)		[below of=sample, yshift=-1.2cm]		{\small Add to dataset};
            \node (train)		[right of=dataset, xshift=2.5cm]		{\small Train ML estimator(s)};
            \node (sample2)		[above of=train]					{\small Sample candidate points};
            \node (evaluate)	[right of=sample, xshift=2.5cm]		{\small Select most interesting};
            \node (end)			[below of=dataset]		{\small Finish sampling};
            \draw[->]			(start) -- (sample);
            \draw[->]			(sample) -- (dataset);
            \draw[->]			(dataset) -- (train);
            \draw[->]			(train) -- (sample2);
            \draw[->]			(sample2) -- (evaluate);
            \draw[->]			(evaluate) -- (sample);
            \draw[->]			(dataset) -- (end);
        \end{tikzpicture}
    \end{center}
	\caption{Diagrammatic representation of active learning. Data is sampled and used to create a data set. This data is used to train an ML estimator (or a committee of estimators), which is used to get an approximation of the labelling on newly sampled data. From this new data the points with the highest uncertainty in their labelling are selected for sampling via the true sampling procedure and added to the data set. This process can be repeated until enough data is collected.}
    \label{fig:active_learning}
\end{figure}

Active learning works as an iterative sampling technique. In this paper we specifically explore a technique called \textit{pool-based sampling}~\cite{settles:2010}, of which a diagrammatic representation can be found in Figure~\ref{fig:active_learning}. In this technique an initial data set is sampled from the parameter space and queried to the labelling procedure (also called the \textit{oracle}). After retrieving the new labels one or more machine learning estimators are trained on the available labelled data. This estimator (or set of estimators) can then provide an approximation of the boundary of the region of interest. We gather a set of candidate (unlabelled) data points, which can for example be sampled randomly or be generated through some simulation, and provide these to the trained estimator. The output of the estimator can then be used to identify which points should be queried to the oracle. For a classification problem this might for example entail finding out which of the candidate points the estimator is most uncertain about. As only these points are queried to the oracle, it will not spend time on evaluating points which are not expected to yield significant information about our region of interest. The selected data points and their labels are then added to the total data set. This procedure -- from creating an estimator to adding new to the data set  -- can be repeated to get an increasingly better estimation of the region of interest and be stopped when e.g.\ the collected data set reaches a certain size or when the performance increase between iterations becomes smaller than a predetermined size.

It should be noted that the active learning procedure as described above has hyperparameters: the size of the initial data set, the size of the pool of candidate data points and the number of candidate data points queried to the oracle in each iteration. The optimal configuration for these parameters is problem dependent and finding them requires a dedicated search. We performed a random search on the hyperparameters of the experiments in Section~\ref{sec:experiments} and selected the best configuration for all experiments. For completeness a discussion on the hyperparameters can be found in Appendix~\ref{app:hyperparams}. We do want to note that in all the active learning configurations we experimented with, active learning always performed at least equally as good as random sampling.

Arguably the most important step in Figure~\ref{fig:active_learning} is to select those points that ought to be queried to the labelling procedure from a large set of candidate data points. As the problems we look at here are classification problems, the closeness to the boundary can be estimated by the uncertainty of the trained estimator on the classification of the model point.

This uncertainty can for example be obtained from an algorithm like Gaussian Processes~\cite{gaussianprocesses}, which has already been successfully applied in high energy physics to aim sampling of new points around 2-dimensional exclusion boundaries~\cite{algp}. Due to the computational complexity of this algorithm it is however limited to low-dimensional parameter spaces, as it scales at best with the number of data points squared~\cite{2018arXiv180911165G}. Because of this, we investigate specifically query-by-committee and query-by-dropout-committee.

\subsection{Query-by-Committee (QBC)}
\label{sec:qbc}
By training multiple machine learning estimators on the same data set, one could use their disagreement on the prediction for a data point as a measure for uncertainty. Data points on which the estimators disagree most are expected to provide the highest information gain. This method is called \textit{query-by-committee} (QBC)~\cite{Seung:1992:QC:130385.130417}. To create and enhance the disagreement among the committee members in uncertain regions the training set can be changed for each estimator (e.g.\ via bagging~\cite{bagging}) or by varying the configuration of the estimator (e.g.\ when using a committee of Neural Networks, each of these could have a different architecture or different initial conditions), such that we get a reasonable amount of diversity in the ensemble.

The disagreement among the estimators can be quantified through metrics like the standard deviation. For binary classification problems it can even be done by taking the mean of the outputs of the set of $N$ estimators. If the classes are encoded as 0 and 1, a mean output of 0.5 would mean maximal uncertainty, so that the uncertainty measure for $N$ estimators could be
\begin{equation}
    \label{eq:rf_uncertainty}
    \textrm{uncertainty} = 1- 2\cdot \left| \frac{1}{N}\sum_{i=1}^N \textrm{prediction}_i - 0.5 \right|.
\end{equation}

The QBC approach is not bound to a specific estimator. If one were to use $N$ estimators of which the training scales linearly with the number of data points $K$, the active learning procedure would have a computational complexity of $\mathcal{O}(NK)$ for each iteration. This allows for the use of large amounts of data, as is needed in high-dimensional parameter space.

\subsection{Query-by-Dropout-Committee (QBDC)}
\label{sec:qmcdc}
Compared to Random Forests, Neural Networks are able to capture more complex data patterns. Using Neural Networks for active learning might therefore be beneficial. However, although query-by-committee can also be used to create a committee of Neural Networks, this is generally ill-advised due to the computational expensive more training of Neural Networks. As an alternative to this, one can also build a committee using Monte Carlo dropout~\cite{2015arXiv150602142G}. This technique uses a Neural Network with dropout layers~\cite{dropout} as estimator. These dropout layers are conventionally used to prevent overtraining (i.e.\ increased performance on the training set at the cost of a reduction in performance on general data sets) by disabling a random selection of the neurons in the preceding layer of the network. This selection is changed at each evaluation, and therefore at each training step of the network. This makes it harder for the network to strengthen a specific node path through the network during training, making it more robust to changes in the input data. At test time the dropout layers are typically disabled and the weights of the layer following the dropout layer are rescaled with a factor of $1 / {(\textrm{1 - dropout fraction})}$, in order to make the network.

In Monte Carlo dropout, on the other hand, the dropout layers are left enabled at test time, making the output of the network vary at each evaluation. Making $x$ predictions on a specific data point can therefore be interpreted as a  set of predictions coming from a committee of $x$ networks. The advantage here however is that only a single network has to be trained. As this method uses Monte Carlo dropout, this method is called Query-by-Dropout-Committee (QBDC)~\cite{dropout_based_al,2015arXiv151106412D, 2018arXiv181103897P}. 

For classification problems one might wonder why not to simply use a softmax output of a normal network (i.e. one not using Monte Carlo dropout) and use this output as a measure for uncertainty. Softmax layers associate a probability for each output class, which could in principle be transformed in an uncertainty. A measure like this can be used the address the uncertainty inherent in the data, also known as the aleatoric uncertainty~\cite{uncertainty-types}. It captures uncertainty around decision boundaries for example, where based on the data the network is insecure on how to classify new data. The output of the softmax layer says however nothing about the uncertainty in the model itself, the epistemic uncertainty~\cite{uncertainty-types}. The epistemic uncertainty on the other hand quantifies the uncertainty related to the model configuration (i.e. that different models would give different predictions).  A single network with a softmax activation on its output would not take epistemic uncertainty into account, as its output has no intrinsic mechanism to capture epistemic uncertainty. This is especially important in regions where training data is sparse. A single network may therefore still give the impression to be very certain in such regions. QBDC, like other ensemble methods, captures model uncertainty by producing more variation in regions where training data is sparse. For active learning, both aleatoric and epistemic uncertainty are important.
\section{Applications in HEP}
\label{sec:experiments}
In this section active learning as a method is investigated using data sets from high energy physics. The experiments investigated here are all classification problems, as these have a clear region of interest: the decision boundary. It should be noted that the methods explored here also hold for regression problems with a region of interest (e.g. when searching for an optimum). Although active learning can also be used to improve the performance of a regression algorithm over the entire parameter space, whether or not this works is highly problem and algorithm dependent, as can for example be seen in Ref.~\cite{Schein2007}.

\subsection{Increase resolution of exclusion boundary}
\label{ssec:application_exclusionboundary}
As there are no significant experimental signals found in ``beyond the standard model'' searches that indicate the presence of unknown physics, the obtained experimental data is used to find the region in the model parameter space that still allowed given the experimental data. Sampling the region around the boundary of this region in high-dimensional spaces is highly non-trivial with conventional methods due to the curse of dimensionality.

We test the application of active learning on a model of new physics with 19 free parameters (the 19-dimensional pMSSM~\cite{Martin:1997ns}). This test is related to earlier work on the generalisation of high-dimensional results, which resulted in SUSY-AI~\cite{Caron:2016hib}. In that work the exclusion information on $\sim 310,000$ model points as determined by the ATLAS collaboration~\cite{Aad2015} was used; the same data is used in this study. We investigate three implementations of active learning: two Random Forest set ups, one with a finite and the other with an infinite pool, and a setup with a QBDC. The performance of each of these is compared to the performance of random sampling, in order to evaluate the added value of active learning. This comparison is quantified by using the following steps:

\begin{enumerate}
    \item Call \texttt{max\_performance} the maximum reached performance for random sampling;
    \item Call $\textrm{N}_{\textrm{random}}$ the number of data points needed for random sampling to reach \texttt{max\_performance};
    \item Call $\textrm{N}_{\textrm{active}}$ the number of data points needed for active learning to reach \texttt{max\_performance};
    \item Calculate the performance gain through
    \begin{equation}
        \label{eq:performancegain}
        \textrm{performance gain} = \frac{\textrm{N}_{\textrm{random}}}{\textrm{N}_{\textrm{active}}}.
    \end{equation}
\end{enumerate}

\noindent The configurations of the experiments were explicitly made identical and were not optimised on their own.

\subsubsection{Random Forest with a finite pool}
\label{ssec:qbc_atlas}
Just as for SUSY-AI we trained a Random Forest classifier on the public ATLAS exclusion data set~\cite{Aad2015} (details on the configuration of this experiment can be found in Appendix~\ref{app:config}). This data set was split into three parts: an initial training set of 1,000 model points, a test set of 100,000 model points and a pool of the remaining $\sim200,000$ model points. As the labeling of the points is 0 for excluded points and 1 for allowed points, after each training iteration the 1,000 new points with their Random Forest prediction closest to 0.5 (following the QBC scheme outlined in Section~\ref{sec:qbc}) are selected from the pool and added to the training set. Using this now expanded dataset a new estimator is trained from scratch. The performance of this algorithm is determined using the test set.

\begin{figure}
    \begin{center}
        \includegraphics[width=0.46\textwidth]{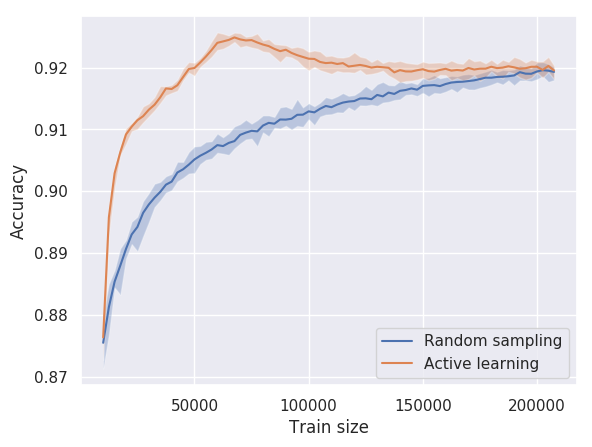}
    \end{center}
    \caption{Accuracy development on model exclusion of the 19-dimensional model for new physics (pMSSM) for random sampling and active learning using a random forest as algorithm and a finite pool. True labeling was provided by ATLAS~\cite{Aad:2015baa}. Active learning quickly starts outperforming the random sampling. The decline in accuracy for active learning, starting from a training size of 60,000, is caused by the limited size of the pool and the fact that the region around the pool is depleted from data around the decision boundary. The bands around the curves show the range in which all curves of that colour lie when the experiment was repeated 7 times.}
    \label{fig:susyai_qbc_atlas}
\end{figure}

This experiment is also performed with all points selected from the pool at random, so that a comparison of the performance of active learning and random sampling becomes possible. The results of both experiments are shown in Figure~\ref{fig:susyai_qbc_atlas}. The bands around the curves in this figure indicate the range in which the curves for 7 independent runs of the experiment lie. The figure shows that active learning outperforms random sampling initially, but after a while random sampling catches up in performance. The decrease in accuracy of the active learning method is caused by an overall lack of training data. After having selected approximately 70,000 points via active learning, new data points are selected further away from this boundary, causing a relative decrease of the weight of the points around the decision boundary, degrading the generalisation performance.

Based on Figure~\ref{fig:susyai_qbc_atlas} the performance gain of active learning over random sampling in the early stages of learning, up to a train size of 50,000, -- as described by Equation~\ref{eq:performancegain} --
lies in the range 3.5 to 4.

\subsubsection{Random Forest with an infinite pool}
\label{ssec:qbc}
A solution to the problem with the finite pool in Section \ref{ssec:qbc_atlas} is to not give it a pool of data but to give it access to a method through which new data points can be generated. In this experiment we create such a method that generates data in the training volume of SUSY-AI with a uniform prior. Although in each iteration only a limited set of candidate points is considered, the fact that this set is sampled anew in each iteration guarantees that the decision boundary is never depleted of new candidate points. Because of this, the pool can be considered infinite. In contrast to the experiment in Section~\ref{ssec:qbc_atlas}, where labeling (i.e., excluded or allowed) was readily available, determining true labeling on these newly sampled data points would be extremely costly. Because of this SUSY-AI~\cite{Caron:2016hib} was used as a stand-in for this labeling process\footnote{Since SUSY-AI has an accuracy of 93.2\% on the decision boundary described by the ATLAS data~\cite{Aad2015}, active learning will not find the decision boundary described by the true labeling in the ATLAS data. However, as the goal of this example is to show that it is possible to find a decision boundary in a high-dimensional parameter space in the first place, we consider this not to be a problem.}. Since we are training a Random Forest estimator, we retrained SUSY-AI as a Neural Network, to make sure the newly trained Random Forest estimator would not be able to exactly match the SUSY-AI model, as this would compromise the possibility to generalise the result beyond this one. The accuracy of this Neural Network was comparable to the accuracy of the original SUSY-AI. Details on the technical implementation can be found in Appendix~\ref{app:config}.

\begin{figure}
    \begin{center}
        \includegraphics[width=0.46\textwidth]{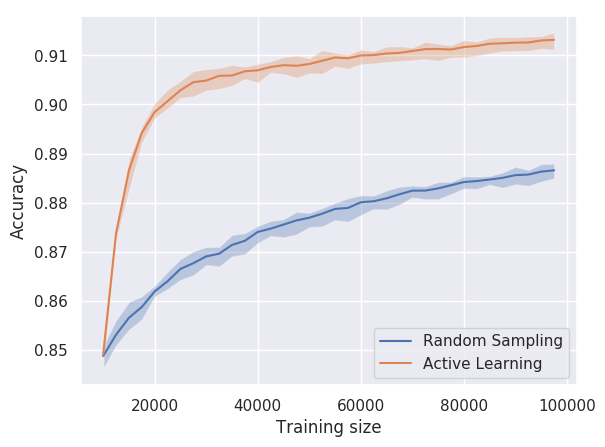}
    \end{center}
    \caption{Accuracy development on model exclusion of the 19-dimensional model for new physics (pMSSM) for random sampling and active learning using a random forest as algorithm and an infinite pool. True labeling was provided by a machine learning algorithm trained on model points and labels provided by ATLAS~\cite{Aad:2015baa}. Here active learning is vastly superior over random sampling, yielding a gain in computational time of a factor of 5 to 6. The bands around the curves show the range in which all curves of that colour lie when the experiment was repeated 7 times.}
    \label{fig:susyai_qbc_random}
\end{figure}

The accuracy development as recorded in this experiment is shown in Figure~\ref{fig:susyai_qbc_random}. The bands again correspond to the ranges of the accuracy as measured over 7 independent runs of the experiment. The gain of active learning with respect to random sampling (as described by Equation~\ref{eq:performancegain}) is 6 to 7. The overall reached accuracy is however lower than in Figure~\ref{fig:susyai_qbc_atlas}, but note that this experiment stopped when a total of $100,000$ points as sampled, compared to the $200,000$ points in the previous experiment.

\subsubsection{QBDC with an infinite pool}
\label{ssec:qbdc}
\begin{figure}
    \begin{center}
        \includegraphics[width=0.48\textwidth]{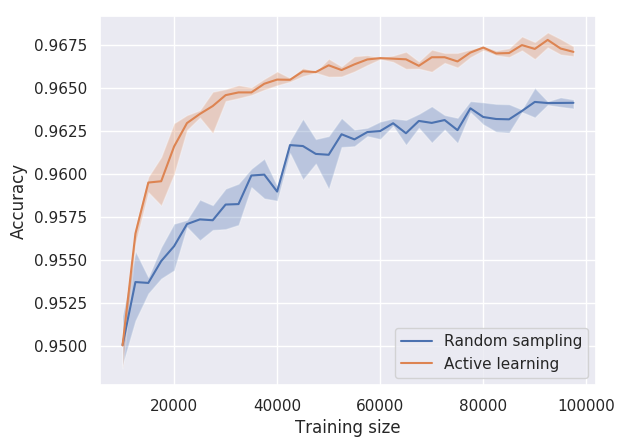}
    \end{center}
    \caption{Accuracy development on model exclusion of the 19-dimensional model for new physics (pMSSM) for random sampling and active learning using a dropout Neural Network with infinite pool. True labeling was provided by a machine learning algorithm trained on model points and labels provided by ATLAS~\cite{Aad:2015baa}. The gain of active learning with respect to random sampling (as described by Equation~\ref{eq:performancegain}) is 3 to 4. The bands show the range in which all curves of that colour lay when the experiment was repeated 7 times.}
    \label{fig:susyai_qbdc}
\end{figure}

To test the performance of QBDC, the infinite pool experiment above was repeated with a QBDC setup. The technical details of the setup can be found in Appendix~\ref{app:config}. The accuracy development plot resulting from the experiment can be seen in Figure~\ref{fig:susyai_qbdc}. The bands around the lines representing the accuracies for active learning and random sampling indicate the minimum and maximum gained accuracy for the corresponding data after running the experiment 7 times. The performance gain (as defined in Equation~\ref{eq:performancegain}) for active learning in this experiment lies in the range 3 to 4.

QBDC sampling is approximately $K$ times faster than ensemble sampling with $K$ committee members for a fixed number of samples, as only one network has to be trained. However, as active learning outperforms random sampling by a factor of 3 to 4, it depends on how expensive training of the estimator is in comparison to how much computational time is gained.

Compared to Figure~\ref{fig:susyai_qbc_atlas} and~\ref{fig:susyai_qbc_random} the accuracies obtained in Figure~\ref{fig:susyai_qbdc} are significantly higher. This can be caused by the fact that the model trained to quantify the performance more strongly resembles the oracle (both of them are Neural Networks with a similar architecture), or that the Neural Network is inherently more capable of capturing the exclusion function.

\subsection{QBC with infinite pool for smaller parameter spaces}
\label{sec:other_dimensionalities}
To investigate the performance of the method on lower dimensional spaces, we perform the QBC experiment of Section~\ref{ssec:qbc} five times. Each of these iterations uses the 19-dimensional parameter space from this original experiment, but in each of these we fix an increasingly larger subset of variables of the original 19. For those familiar with SUSY: we consecutively fix the variables associated with the slepton sector, the electroweak (EW) sector, the higgs sector and the third generation. An overview of the variables and their fixed values can be found in  Table~\ref{tab:reduced_dimensionality_setup}. The specific values to which the variables were fixed were determined using SUSY-AI: we required that whatever value selected, there was a balance in the fraction of allowed and excluded points. To make results from these experiments comparable to the experiments in Section \ref{ssec:qbc}, each of the active learning parameters (size of the initialisation set, iteration size, sample size and maximum size) is scaled down by a factor

\begin{equation}
    \textrm{scaling} = \left( \frac{\textrm{number of free parameters}}{19}\right)^2
\end{equation}

\noindent This avoids, amongst other things, for example that the initial data set is too large for active learning to be most effective.

\begin{table}[]
    \centering
    \caption{List of parameters of the 19-dimensional pMSSM in the reduced dimensionality experiments. Parameters with a fixed value have this value denoted in the respective column. Parameters that are left free are indicated by a dash.}
    \begin{tabular}{lccccc}
        \hline
        \hline
        \textbf{\# Fixed} & \textbf{5} & \textbf{9} & \textbf{11} & \textbf{15} & \textbf{18}  \\
        
        \hline
        \texttt{M1} (GeV)        & - & 1750 & 1750 & 1750 & 1750 \\
        \texttt{M2} (GeV)       & - & 1750 & 1750 & 1750 & 1750 \\
        \texttt{M3} (GeV)        & - & - & - & - & - \\
        \texttt{mL1} (GeV)       & 1750 & 1750 & 1750 & 1750 & 1750 \\
        \texttt{mL3} (GeV)       & 1750 & 1750 & 1750 & 1750 & 1750 \\
        \texttt{mE1} (GeV)       & 1750 & 1750 & 1750 & 1750 & 1750 \\
        \texttt{mE3} (GeV)       & 1750 & 1750 & 1750 & 1750 & 1750 \\
        \texttt{mQ1} (GeV)       &   & - & - & - & 1750\\
        \texttt{mQ3} (GeV)       & - & - & - & 1750 & 1750 \\
        \texttt{mU1} (GeV)       & - & - & - & - & 1500 \\
        \texttt{mU3} (GeV)       & - & - & - & 3000 & 3000 \\
        \texttt{mD1} (GeV)       & - & - & - & - & 2000 \\
        \texttt{mD3} (GeV)       & - & - & - & 2000 & 2000 \\
        \texttt{At}        & - & - & 3200 & 3200 & 3200 \\
        \texttt{Ab}        & - & - & - & 2000 & 2000 \\
        \texttt{Atau}      & 2000 & 2000 & 2000 & 2000 & 2000 \\
        \texttt{mu} (GeV)        & - & 200 & 200 & 200 & 200\\
        \texttt{mA2} (GeV$^2$)       & - & - & 10$^7$ & 10$^7$ & 10$^7$ \\
        \texttt{tan(beta)} & - & 10 & 10 & 10 & 10 \\
        \hline
        \hline
    \end{tabular}
    \label{tab:reduced_dimensionality_setup}
\end{table}

\begin{figure}
    \begin{center}
        \includegraphics[width=0.48\textwidth]{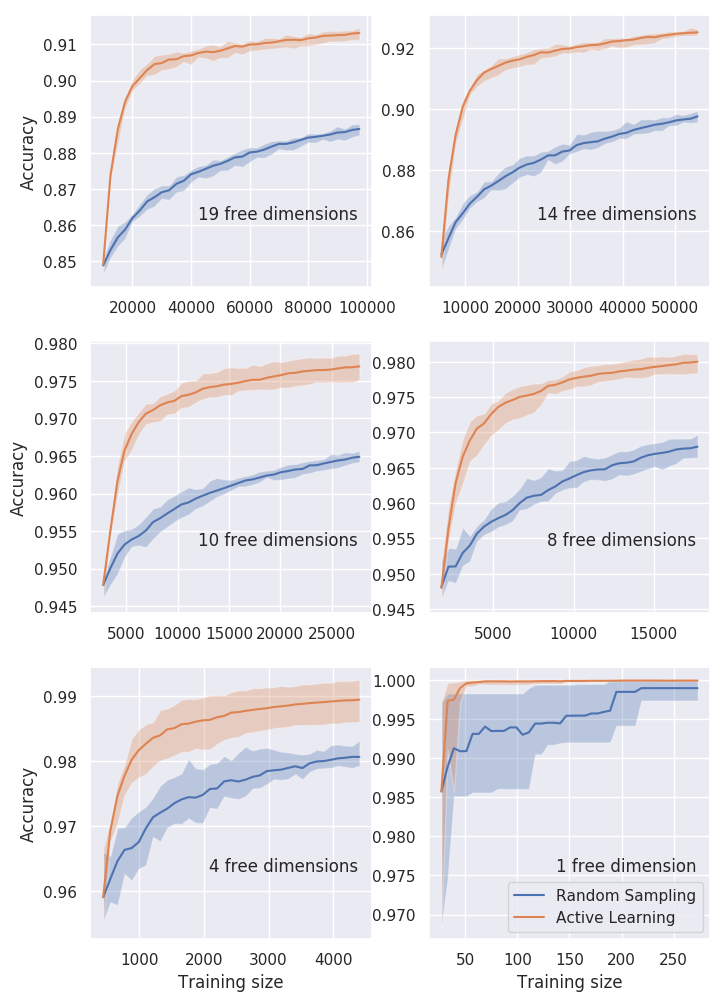}
    \end{center}
    \caption{Accuracy development on model exclusion in the simplified versions of the 19-dimensional model for new physics (pMSSM) for random sampling and active learning using a random forest as algorithm and an infinite pool. True labeling was provided by a machine learning algorithm trained on model points and labels provided by ATLAS~\cite{Aad:2015baa}. The bands show the range in which all curves of that colour lay when the experiment was repeated 7 times. The subsplots show the development for different dimensionalities for the simplified models. The relative gain of active learning over random sampling is approximately equal for all six of these plots, showing that active learning has added benefit even for smaller parameter spaces. }
    \label{fig:susyai_qbc_random_reduceddim}
\end{figure}

The results of the five reduced dimensionality experiments can be found in Figure \ref{fig:susyai_qbc_random_reduceddim}. We have also included the results of Section \ref{ssec:qbc} for comparison. The gain over active learning with respect to random sampling (Equation \ref{eq:performancegain}) differs slightly from plot to plot and can be found in the last column of Table \ref{tab:time}. From this table it can be seen that active learning is of added value, even when applied to parameter spaces with a smaller dimensionality.

However, although the relative gain in performance is an informative measure, when deciding to apply active learning one also has to consider the additional computational overhead that the procedure adds to the sampling process. How large of an overhead one is willing to accept depends largely on the cost associated with querying the oracle. If a query to the oracle would be very expensive (say $\mathcal{O}(\textrm{day})$) an overhead of an hour may be completely worth the wait, whereas such an overhead would be unreasonable for inexpensive oracles.

To investigate this influence of oracle query times we kept track of the time taken by the sampling process and oracle query times for active learning in this and the previous section. Using this information we calculated the time needed for active learning to obtain the maximum performance by random sampling. Comparing this time to the time needed to randomly sample the necessary amount of points for this performance gives a time-based comparison between the two sampling methods. In Table~\ref{tab:time} the following metric is used

\begin{equation}
    \label{eq:timegain}
    \textrm{time metric} = \frac{\textrm{time taken by random sampling}}{\textrm{time taken by active learning}}.
\end{equation}

\noindent The table shows the results for this metric for the experiments in this and the previous section, averaged out over the seven performed iterations. From these results it can be concluded that active learning can provide a benefit over random sampling even for low dimensional parameter spaces, as long as the oracle query time is long enough. An alternative interpretation could be that if one would be able to find an invertible transformation that projects the data to a lower dimensional space, active learning might not be necessary -- as long as the oracle query time is short enough.

An important note should be made about the exact oracle query times in Table~\ref{tab:time} however. Although the ratios reported in the table are independent on the used hardware and the used active learning method, this is not the case of the dependence of these ratios on the quoted oracle query times. Where the time taken by random sampling is only dependent on this oracle query time and has negligible overhead, this is not the case for active learning. So if the oracle query time stays the same but the computer becomes for example twice as slow (or the machine learning method would be replaced with a method that is twice as expensive), the ratio would drop with a factor anywhere between 1 and 2, the exact value depending on the oracle query time and the time taken by the active learning overhead. The general trend that active learning can be beneficial for lower dimensional parameter spaces if the oracle time is just long enough -- or alternatively: that active learning becomes more and more relevant as one considers higher and higher dimensional parameter spaces -- is however unaffected by this hardware dependence.

\begin{table*}[t]
    \centering
    \caption{Relative time gain (Equation \ref{eq:timegain}) of the active learning procedure with respect to random sampling when attempting to obtain the maximum performance obtained with random sampling. The table shows this gain for different parameter space dimensionalities (rows) and for different costs for querying the oracle (columns). The shown results are the averages over all seven iterations of the experiment in Section~\ref{sec:other_dimensionalities}. The last column shows the performance gain (Equation \ref{eq:performancegain}), irrespective of the active learning overhead. Noticeable is that even for lower dimensional parameter spaces active learning can be beneficial, namely when the computational cost associated with querying the oracle becomes high enough.}
    \begin{tabular}{l|cccccccccc|c}
    \hline
    \hline
    \textbf{\# free parameters} & \textbf{1$\mu$s} & \textbf{10$\mu$s} & \textbf{100$\mu$s} & \textbf{1ms} & \textbf{10ms} & \textbf{100ms} & \textbf{1s} & \textbf{10s} & \textbf{100s} & \textbf{1000s} & \textbf{$\lim_{t \to \infty}$}\\
    \hline
    19 & 0.0112 & 0.1101 & 0.9674 & 4.3736 & 6.7504 & 7.1383 & 7.1796 & 7.1837 & 7.1842 & 7.1842 & 7.1842\\
    14 & 0.0068 & 0.0672 & 0.6149 & 3.3239 & 5.9413 & 6.4491 & 6.5047 & 6.5103 & 6.5109 & 6.511 & 6.511\\
    10 & 0.0046 & 0.0456 & 0.4282 & 2.6664 & 5.587 & 6.2742 & 6.3524 & 6.3603 & 6.3611 & 6.3612 & 6.3612\\
    8 & 0.0021 & 0.0213 & 0.2058 & 1.5164 & 4.1751 & 5.0627 & 5.1727 & 5.1839 & 5.1851 & 5.1852 & 5.1852\\
    4 & 0.0005 & 0.0053 & 0.0525 & 0.4784 & 2.5256 & 4.4148 & 4.7718 & 4.8107 & 4.8146 & 4.815 & 4.815\\
    1 & <0.0001 & 0.0001 & 0.0012 & 0.012 & 0.1152 & 0.8196 & 2.1106 & 2.5051 & 2.5528 & 2.5577 & 2.5582\\
    \hline
    \hline
    \end{tabular}
    \label{tab:time}
\end{table*}

\subsection{Identifying uncertain regions and steering new searches}
\label{ssec:application_steeringsearches}
Instead of using active learning e.g. to iteratively increase the resolution on a decision boundary, the identification of uncertain regions of the parameter space on which active learning is built can also be used to identify regions of interest.

\begin{figure}
    \begin{center}
        \includegraphics[width=0.46\textwidth]{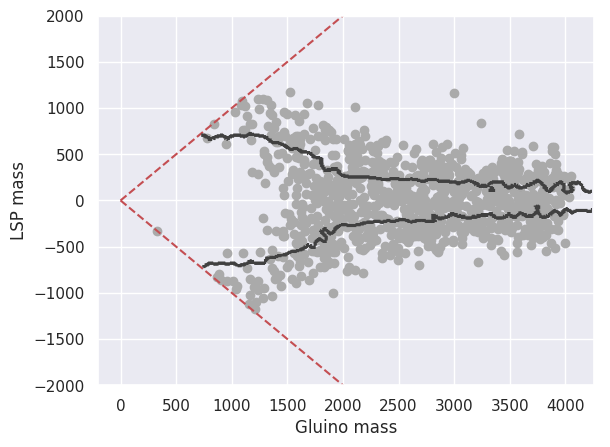}
    \end{center}
    \caption{The model points that were selected in a pool-based sampling projected on the gluino mass ($m_{\tilde{g}}$) - LSP mass ($m_{\tilde{\chi}^0_1}$) plane. The algorithm did not have direct access to the variables on the axes, but was nevertheless able to sample points in the region around the decision boundary, indicated by the solid black line in the figure. The dashed red lines indicate the boundary of the model, outside of which no data points will be sampled (caused by the fact that no supersymmetric particle can be lighter than the LSP).}
    \label{fig:steeringsearches}
\end{figure}

For example, in high energy physics one could attempt to identify model points around the exclusion boundary in a high-dimensional model. These model points could then be used as targets for new searches or even new experiments. This is an advantage over the conventional method of trying to optimise a 2-dimensional projection of the exclusion region, as this method works over the full dimensionality of the model, which thereby can respect a more detailed account of the underlying theory that is being tested for. One could even go a step further by reusing the same pool for these search-improvement studies, so that regions of parameter space that no search has been able to exclude can be identified. Analogous to this one could also apply this method to find targets for the design of a new experiment. 

To test the application of this technique in the context of searches for new physics we trained a Neural Network on the publicly available ATLAS exclusion data on the pMSSM-19~\cite{Aad2015}, enhanced with the 13\,TeV exclusion information as calculated by~\cite{Barr:2016inz}. The technical setup is detailed in Appendix~\ref{app:config}. We sampled $\sim87,000$ model points in the SUSY-AI parameter space~\cite{Caron:2016hib} using \texttt{SOFTSUSY 4.1.0}~\cite{softsusy} as spectrum generator and selected 1,000 points with the highest uncertainty following the QBDC technique outlined in Section~\ref{ssec:qbdc}.

Figure~\ref{fig:steeringsearches} shows the sampled model points in the gluino mass - LSP mass projection. As the LSP mass was not directly one of the input parameters, the fact that the selected points are nevertheless well-sampled in the region of the decision boundary, we conclude that the active learning algorithm did successfully find the decision boundary in the 19-dimensional model.

We conclude this section by noting that in all the active learning experiments in this section new points were selected exclusively with active learning. In more realistic scenarios the user can of course use a combination of random sampling and active learning, in order not to miss any features in parameter space that were either unexpected or not sampled by the initial dataset.
\section{Conclusion}
\label{sec:conclusion}

In this paper we illustrated the possibility to improve the resolution of regions of interest in high-dimensional parameter spaces. We specifically investigated query-by-committee and query-by-dropout-committee as tools to constrain parameters and the possibility to improve the identification of uncertain regions in parameter space to steer the design of new searches. We find that all active learning strategies presented in this paper query the oracle more efficiently than random sampling, by up to a factor of 7. By reducing the dimensionality of the  19-dimensional parameter space used in the experiments, we have additionally shown that active learning can be beneficial even for parameter spaces with a lower dimensionality.

One of the limiting factors of the techniques as presented in this paper is the fact that a pool of candidate points needs to be sampled from the parameter space. If sampling candidate points randomly yields too few interesting points, generative models can be used to sample candidate points more specifically.

The code for all performed experiments is made public on GitHub \href{https://github.com/bstienen/active-learning}{\faGithub}\footnote{\url{https://github.com/bstienen/active-learning}}.
\section*{Acknowledgements}
This research is supported by the Netherlands eScience Center grant for the \textit{iDark: The intelligent Dark Matter Survey} project.

\newpage
\begin{appendices}
\section{Active learning hyperparameters}
\label{app:hyperparams}
The active learning procedure as implemented for this paper has three hyperparameters:
\begin{itemize}
    \item[-] \texttt{size\_initial}: The size of the data set used at the start of the active learning procedure;
    \item[-] \texttt{size\_sample}: The size of the pool of candidate data points to be sampled in each iteration
    \item[-] \texttt{size\_select}: The number of data points to select from the pool of candidate data points and query to the oracle.
\end{itemize}

Which settings are optimal depends on the problem at hand, although some general statements can be made about the possible values for these hyperparameters. To illustrate this we performed a hyperparameter optimisation for the experiment in Section~\ref{ssec:qbc}, although it should be noted that this optimisation was performed only for illustration purposes and was not used to configure the experiments in this paper.

The \texttt{size\_initial} for example configures how well the first trained machine learning estimator approximates the oracle. If this approximation is bad, the first few sampling iterations will sample points in what will later turn out to be uninteresting regions. A higher value for \texttt{size\_initial} would therefore be preferable over a smaller value, although this could diminish the initial motivation for active learning: avoiding having to run the oracle on points that are not interesting with respect to a specific goal.

\begin{figure}
    \centering
    \includegraphics[width=0.46\textwidth]{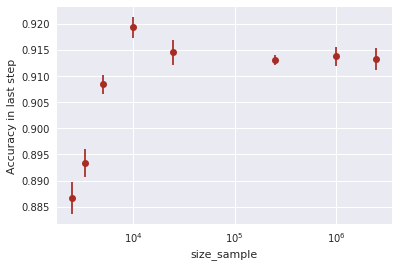}
    \caption{The dependence of the accuracy in the last iteration of the active learning procedure on the number of candidates in each iteration. The error bars indicate the range within which the accuracies over 7 runs lie. As described in the text, an optimum value can be observed, although it should be noted that this value also depends on the number of data points selected in each iteration.}
    \label{fig:hyperparam_size_sampled}
\end{figure}

The \texttt{size\_sample} parameter however will have an optimum: if chosen too small the selected samples will be more spread out and possibly less interesting points will be queried to the oracle. If chosen too high on the other hand the data could be focused in a specific subset of the region of interest because the trained estimator happens to have a local minimum there. The existence of an optimal value for \texttt{size\_sample} can be seen in Figure~\ref{fig:hyperparam_size_sampled}.

\begin{figure}
    \begin{center}
        \includegraphics[width=0.46\textwidth]{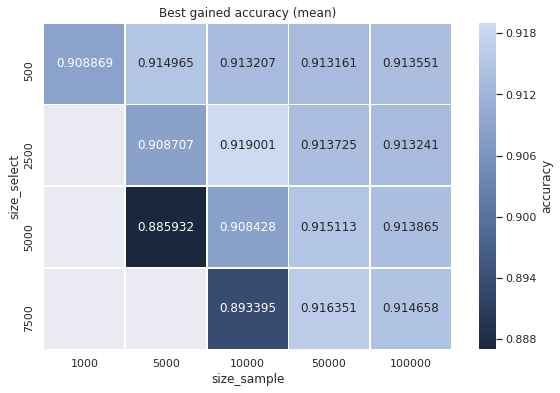}
    \end{center}
    \caption{The dependence of the accuracy in the last iteration of the active learning procedure on the number of candidates and the number of points selected for querying to the oracle in each iteration. The last iteration was defined as the last iteration before 100,000 data points were selected, meaning that a setup with \texttt{size\_select} equal to 500 had more iterations than a setup with a \texttt{size\_select} of 7500 for example.}
    \label{fig:hyperparam_grid}
\end{figure}

It should be noted that the location of the optimum does not only depend on \texttt{size\_sample}, but also on \texttt{size\_select}. If one were to set \texttt{size\_select} to 1, the size of the candidate pool is best as large as possible, in order to be sure that the selected point is really the most informative one you can select. This would avoid the selection of clustered data points, but this comes at the cost of having to run the procedure for more iterations in order to get the same size for the final data set. This would however be very expensive if the cost for training the ML estimator(s) is very high. The dependence of the accuracy on these two variables is shown in Figure~\ref{fig:hyperparam_grid}, in which the accuracy gained in the last step of the active learning procedure is shown for different configurations of these two parameters. The script to generate this figure can be found on GitHub \href{https://github.com/bstienen/active-learning}{\faGithub}.

\section{Experiment configuration}
\label{app:config}
All networks were trained using Keras~\cite{keras} with a Tensorflow~\cite{tensorflow} backend linked to CUDA~\cite{cuda}. scikit-learn~\cite{scikit-learn} was used when implementations of RandomForest were used.

\subsection*{Increase resolution of exclusion boundary}
The configuration of the active learning procedure can be found in Table~\ref{tab:al_configuration}. The experiments are denoted by the section in this paper in which they were covered.

\begin{table}[]
    \centering
    \caption{Configuration for the active learning procedures in Section \ref{ssec:application_exclusionboundary}.}
    \begin{tabular}{l|c|c|c}
        \hline
        \hline 
         & \textbf{III.A.1} & \textbf{III.A.2} & \textbf{III.A.3} \\
        \hline
        Initial dataset & \multicolumn{3}{c}{10,000} \\
        \hline
        Step size & \multicolumn{3}{c}{2,500} \\
        \hline
        \#candidates & remaining pool & \multicolumn{2}{c}{100,000} \\
        \hline
        Maximum size & until pool empty & \multicolumn{2}{c}{100,000}\\
        \hline
        Committee size & \multicolumn{2}{c|}{100} & 25 \\
        \hline
        \#iterations & \multicolumn{3}{c}{7} \\
        \hline
        \#test points & \multicolumn{3}{c}{1,000,000} \\
        \hline
        \hline
    \end{tabular}
    \label{tab:al_configuration}
\end{table} 

\vspace{0.3cm}
\paragraph*{\textbf{Random Forest with a finite pool}}\hspace{0.5cm}
The trained Random Forest classifier followed the defaults of scikit-learn~\cite{scikit-learn}: it consisted out of 10 decision trees with gini impurity as splitting criterion.
\begin{table}
    \centering
    \caption{Network architecture for the oracle in the ``Random Forest with an infinite pool'' and  the ``QBDC with an infinite pool'' experiments.}
    \begin{tabular}{lccr}
        \hline
        \hline
        \textbf{Layer type} & \textbf{Config.} & \textbf{Output shape} & \textbf{Param. \#} \\
        \hline
        Input & & (None,19) & 0 \\
        Dense & 500 nodes & (None, 500) & 10,000 \\
        Activation & selu & (None, 500) & 0 \\
        Dense & 100 nodes & (None, 100) & 50,100 \\
        Activation & selu & (None, 100) & 0 \\
        Dense & 100 nodes & (None, 100) & 10,100 \\
        Activation & selu & (None, 100) & 0 \\
        Dense & 50 nodes & (None, 50) & 5,050 \\
        Activation & selu & (None, 50) & 0 \\
        Dense & 2 nodes & (None, 2) & 102 \\
        Activation & softmax & (None, 2) & 0 \\
        \hline
        Total params: & & & 75,352 \\
        \hline
        \hline
    \end{tabular}
    \label{tab:architecture_qbc}
\end{table}

\vspace{0.3cm}
\paragraph*{\textbf{Random Forest with an infinite pool}}\hspace{0.5cm}
For active learning we trained a Random Forest~\cite{randomforest} classifier that consisted out of 100 decision trees with gini impurity as splitting criterion. All other settings were left at their default values.

As the oracle we used a neural network with the architecture in Table~\ref{tab:architecture_qbc}. This network was optimised using Adam~\cite{ADAM} on the binary cross entropy loss. The network was trained using the ATLAS pMSSM-19 dataset~\cite{Aad:2015baa} for 300 epochs with the EarlyStopping~\cite{EarlyStopping} callback using a patience of 50.

\vspace{0.3cm}
\paragraph*{\textbf{QBDC with an infinite pool}}\hspace{0.5cm}
The network architecture for the trained neural network used for active learning can be found in Table~\ref{tab:architecture_qbdc}. The active learning network was optimized using Adam~\cite{ADAM} on a binary cross-entropy loss. It was fitted on the data in 1000 epochs, a batch size of 1000 and the EarlyStopping~\cite{EarlyStopping} callback using a patience of 20. The neural network from the infinite pool experiment described above is also used in this experiment.\vspace{0.3cm}

\begin{table}
    \centering
    \caption{Network architecture for the ``QBDC with an infinite pool'' experiment.}
    \begin{tabular}{lccr}
        \hline
        \hline
        \textbf{Layer type} & \textbf{Config.} & \textbf{Output shape} & \textbf{Param. \#} \\
        \hline
        Input & & (None,19) & 0 \\
        Dense & 500 nodes & (None, 500) & 10,000 \\
        Activation & relu & (None, 500) & 0 \\
        Dropout & 0.2 & (None, 500) & 0 \\
        Dense & 100 nodes & (None, 100) & 50,100 \\
        Activation & relu & (None, 100) & 0 \\
        Dropout & 0.2 & (None, 100) & 0 \\
        Dense & 100 nodes & (None, 100) & 10,100 \\
        Activation & relu & (None, 100) & 0 \\
        Dropout & 0.2 & (None, 100) & 0 \\
        Dense & 50 nodes & (None, 50) & 5,050 \\
        Activation & relu & (None, 50) & 0 \\
        Dropout & 0.2 & (None, 50) & 0 \\
        Dense & 2 nodes & (None, 2) & 102 \\
        Activation & softmax & (None, 2) & 0 \\
        \hline
        Total params: & & & 75,352 \\
        \hline
        \hline
    \end{tabular}
    \label{tab:architecture_qbdc}
\end{table}

\subsection*{Identifying uncertain regions and steering new searches}
The network architecture for the trained neural network can be found in Table~\ref{tab:architecture_qbdc}. The network was optimized using Adam~\cite{ADAM} on a binary cross entropy loss. It was fitted on the data in 1000 epochs, a batch size of 1000 and with the EarlyStopping~\cite{EarlyStopping} callback using a patience of 50..

The network was trained on the z-score normalised ATLAS dataset~\cite{Aad:2015baa} of 310,324 data points, of which 10\,\% was used for validation. 
\end{appendices}

\bibliographystyle{plain}
\bibliography{refs}

\end{document}